\documentclass[a4paper,english]{rnti}

\usepackage[T1]{fontenc}
\usepackage[utf8]{inputenc}
\usepackage{hyperref}

\usepackage{url}

\usepackage[table]{xcolor}
\usepackage{booktabs}
\usepackage{graphicx}
\usepackage{multirow}
\makeatletter
\newcommand\foottotoref[1]{\protected@xdef\@thefnmark{\ref{#1}}\@footnotemark}
\makeatother

\titrecourt{A Benchmark for Toxic Comment Classification on Civil Comments Dataset}

\nomcourt{C. Duchêne et al.}

\titre{A benchmark for toxic comment classification on Civil Comments dataset}

\auteur{Corentin Duchêne,
        Henri Jamet,
        Pierre Guillaume,
        Réda Dehak}

\affiliation{
    \texttt{firstname.lastname@epita.fr}\\
    EPITA Speaker and Language Recognition Group (ESLR),\\    
    Laboratoire de Recherche de l'EPITA (LRE), France%
 }

\resume{%
Toxic comment detection on social media has proven to be essential for content moderation. This paper compares a wide set of different models on a highly skewed multi-label hate speech dataset. We consider inference time and several metrics to measure performance and bias in our comparison. We show that all BERTs have similar performance regardless of the size, optimizations or language used to pre-train the models. RNNs are much faster at inference than any of the BERT. BiLSTM remains a good compromise between performance and inference time. RoBERTa with Focal Loss offers the best performance on biases and AUROC. However, DistilBERT combines both good AUROC and a low inference time. All models are affected by the bias of associating identities. BERT, RNN, and XLNet are less sensitive than the CNN and Compact Convolutional Transformers.
}

\summary{%
La détection des commentaires toxiques sur les réseaux sociaux s'est avérée essentielle pour la modération de contenu. Dans cet article, nous comparons un large éventail de modèles sur un ensemble de données multilabels de discours haineux fortement biaisé. Nous prenons en compte dans notre comparaison le temps d'inférence, les performances et le biais à travers différentes métriques. Nous avons découvert que tous les BERT ont des performances similaires, indépendamment de leur taille, des optimisations ou du langage utilisé pour le pré-entraînement. BiLSTM reste un bon compromis entre la performance et le temps d'inférence. RoBERTa avec Focal Loss demeure le moins biaisé de tous. DistilBERT a le temps d'inférence le plus faible des BERT. Enfin, tous les modèles sont affectés par le biais d'association des identités à la toxicité. BERT, RNN et XLNet y sont moins sensibles que les CNN et les Compact Convolutional Transformers.
}

\begin{document}

%
\section{Introduction}

Toxic comment detection on social media has proven to be essential for content moderation. According to the French Minister of Education, 18\% of French students were victims of harassment on social networks in 2021. At the same time, the number of posts on these platforms has been increasing. In 12 years, the number of tweets per day has increased tenfold to reach 500 million today\footnote{\href{https://www.internetlivestats.com/twitter-statistics/}{Twitter usage statistics - Internet Live Stats}}.

This shows that the rapid and targeted detection of toxic comments on social networks became a crucial issue in ensuring the cohesion of society. Therefore, this can only be done by automating online moderation.

Nowadays, the types of models performing well on text classification and representing state-of-the-art are transformer-based models \citep{32} such as BERT \citep{33}. \citet{34} refined a pre-trained BERT model for identifying offensive language, automatically categorizing hate types, and identifying the target of the comment. \citet{35} used BERT for training and testing cross-data sets, \citet{36} separately refined BERT on several datasets for hate speech and offensive language detection.

In this study, we compare state-of-the-art models in natural language processing, such as BERT, and in vision applied to text, such as ResNet and Vision Transformers. To our best knowledge, we did not find in the state-of-the-art a detailed comparison of all these models on a wide range of metrics using the same training conditions and same training and testing datasets. As never before,  the same methodology and dataset are used throughout our analysis to focus on performance, bias measurement, and inference time. We have tuned each of our models to achieve the best performance. This work's result should help determine which model can be used in practice.

Our comparison was performed using the same training and testing datasets extracted from the Civil Comments 2019\footnote{\href{https://www.kaggle.com/c/jigsaw-unintended-bias-in-toxicity-classification/data}{Jigsaw unintended bias in toxicity classification Kaggle}\label{civil-comments-data}}. This dataset is a multi-label dataset with imbalanced classes provided by Jigsaw/Conversation AI. For this dataset, we know the targeted identity for some comments, so that we can evaluate the biases during classification.

The rest of the paper is organized as follows: Section \ref{sec:methodology} describes the dataset, and the models used in the comparison. The experiment protocol and the results' analysis are presented in sections~\ref{sec:experiments} and~\ref{sec:result}. Finally, section \ref{sec:conclusion} concludes the paper.

\section{Methodology}
\label{sec:methodology}

\subsection{Dataset}

In 2017, the comment-hosting platform Civil Comments closed. The company has made its 1.8 million comments public to support  research to understand and improve civility detection in online conversations. The Jigsaw team  supports this action; each comment was shown to 10 annotators and asked them to “Rate the toxicity of the comment”. To ensure the accuracy of the ratings, some comments were seen by more than 100 annotators. For all comments, the value obtained at the end for each class is the fraction of positive annotations over the number of annotators. All comments were classified into seven categories: \texttt{toxicity}, \texttt{severe\_toxicity}, \texttt{obscene}, \texttt{threat}, \texttt{insult}, \texttt{indentity\_attack}, and \texttt{sexual\_explicit}.

\begin{table}[ht]
\centering
{
\footnotesize
\begin{tabular}{lp{4cm}}
\hline
{\bf Category} & {\bf Identity Options}\\\hline
Gender & Male, Female, Transgender, Other gender \\\hline
Sexual Orientation & Heterosexual, Homosexual, Bisexual, Other sexual orientation \\\hline
Religion & Christian, Jewish, Muslim, Hindu, Buddhist, Atheist, Other religion \\\hline
Race or ethnicity & Black, White, Latino, Other race or ethnicity\\\hline
Disability & Physical disability, Intellectual or learning disability. Psychiatric disability or mental illness, Other disability\\\hline
\end{tabular}
}
\caption{List of identity options presented to the annotators.}
\label{tab:listidentity}
\end{table}

\begin{table}[ht]
\centering
\begin{tabular}{lcc}
\hline
{\bf Subgroup} & {\bf Count} & {\bf Percent Toxic} \\\hline
all comments & 1 999 516 & 7.99\% \\\hline
male & 48 870 & 15.05\% \\
female & 58 584 & 13.66\% \\
transgender & 2 759 & 21.13\% \\
heterosexual & 1 432 & 22.56\% \\
homosexual & 12 062 & 28.28\% \\\hline
\end{tabular}
\caption{Percentage of comments labeled toxic for a selection of identities.}
\label{tab:peridentity}
\end{table}

In addition, a subset of 450,000 examples from this dataset was tagged with identity (Table~\ref{tab:listidentity}) using a list of questions, such as "What genders are referenced in the comment?" or "What races or ethnicities are referenced in the comment?”. Again, the score obtained for each identity class is the fraction of annotators who mentioned the identity out of the number of evaluators. We can see in Table~\ref{tab:peridentity} that there is an imbalance in toxicity percentage annotation between different identities. 

\subsection{Preprocessing}

The comments are labeled with the probabilities of belonging to a class (for all toxicity and identity classes). To determine whether a comment is considered positive or negative for a class, we applied a threshold: if the probability is greater than 0.5; we assume that the comment is positive for that class, otherwise negative.

We notice that the classes are highly unbalanced. The label \texttt{severe\_toxicity} is rarely activated on the whole dataset, as shown in Table \ref{tab:counthatetype}. For this reason, this class has been removed from the classes to be predicted to limit the number of classes to six.

\begin{table}[ht]
\centering
\begin{tabular}{lr}
\hline
{\bf Hate subtype} & {\bf Count} \\\hline
toxicity & 159 782 \\
severe\_toxicity & 13 \\
obscene & 10 671 \\
sexual\_explicit & 5 127 \\
identity\_attack & 14 761 \\
insult & 118 079 \\
threat & 4 725 \\
\hline
\end{tabular}
\caption{Count of comments for each subtype of hate speech.}
\label{tab:counthatetype}
\end{table}

The following transformations are applied to each comment:
\begin{itemize}
    \item Remove HTML tags
    \item Remove URL
    \item Remove diacritics
    \item transform to lowercase
    \item Remove white space
    \item Remove NA or empty
\end{itemize}

The dataset available on Kaggle\foottotoref{civil-comments-data} already provided a split into train and test subsets. It is assumed that the distributions of labels and subgroups between the two subsets are similar but not exact.

To deal with the problem of unbalanced dataset, during the training step, we re-balance the toxicity classes. To do this, we will apply a negative down-sampling: we keep only 10\% of the randomly chosen examples without toxicity (all 6 toxicity classes not enabled), and we keep all the examples with at least one of the 6 classes enabled. In fine, there are as many examples with all the negative classes as there are examples with at least one positive class. In total, the size of the training set is 310 000 examples. It is important to note that no re-balancing is done on the test subset.

\subsection{Models}

Most of the trained transformers are based on BERT,
it stands for Bidirectional Encoder Representations from Transformers, Google developed it in 2018. It is a Transformer-based model that only uses the encoder part of the Transformer. BERT model can also be used for classification. It uses a specific token \texttt{<CLS>} in the
beginning of each sequence for classification purposes. 
In our comparison, we assume BERT as our baseline model. 

Despite the excellent results in different benchmarks, this is a model that has some limitations. 
Since the release of BERT, different models were proposed to address some BERT
limitations. For this reason, we will investigate the performance of recent transformer language models: DistilBERT \citep{39}, AlBERT \citep{38}, RoBERTa \citep{37}, XLM RoBERTa \citep{41}, BERTweet \citep{42}, HateBERT \citep{40}, XLNet \citep{43} and Compact Convolutional Transformer (CCT) \citep{44}.

\subparagraph{DistilBERT} was proposed by \citet{39}. It is a distilled \citep{distill} version of the BERT model. The new model has 40\% less parameters, runs 60\% faster while preserving over 95\% of BERT’s performances.

\subparagraph{AlBERT} \citep{38}, which stands for “A Lite BERT”, was made available in an open source version by Google in 2019. The model was built with the original BERT structure, but designed to drastically reduce parameters (by 89\%) using sharing parameters across the hidden layers of the network, and factorizing the embedding layer. All of this was accomplished with an accuracy reduction of 82.3\% to 80.1\% on average over a list of datasets.

\subparagraph{RoBERTa} \citep{37} is a modification of BERT model proposed by Facebook AI in 2019. To improve end-task performance, Roberta uses a byte-level Byte-Pair-Encoding \citep{bytepair_encoding} as a tokenizer. Hence, the tokenizer contains more than 50k words, which increases the embedding layer size and the number of learning parameters. Regarding prior training, Roberta has been trained on Masked Language Modeling (MLM) and Causal Language Modeling (CLM). In causal language modeling, the model tries to predict masked token with only the left or right tokens in the sentence, which makes the prediction unidirectional.

\subparagraph{XLM RoBERTa} \citep{41} is a multilingual version of RoBERTa. It is pre-trained on 2.5 TB of filtered CommonCrawl data containing 100 languages.

\subparagraph{BERTweet} \citep{42} is a BERT-based model trained on a huge English tweet corpus proposed by Nvidia. It was trained using the Roberta procedure on language modeling task. The corpus used for the training is about 820 millions (80Go) of English tweets. In order to train the model, huge capacity resources were needed: 8 v100 GPUs with 32 GB each. BERTweet has showed to outperform Roberta base model on the following tweets task: Part-of-speech tagging, Named-entity recognition and text classification.

\subparagraph{HateBERT} \citep{40} is a model published in Association for Computational Linguistics conference. It uses a pretrained BERT base model. This model has been fine-tuned for a language modeling task on specific social network dataset RAL-E (Reddit Abusive Language English dataset). The dataset is made of 1 492 740 different sentences from Reddit and contains hate speech, offensive and abusive phrases. The model has also been fine-tuned on 3 different datasets: OfensEval, AbusEval and HatEval beating the state of the art on these 3 datasets.

\subparagraph{XLNet} \citep{43} is a large bidirectional transformer that uses improved
training methodology, larger training dataset, and more
computational power. XLNet outperformed BERT on 20 tasks, such as question
answering, natural language inference, sentiment
analysis, etc.\\

For all these models, we concatenate the output of the last 4 layers in one large features vector and stacked two dense layers to get a vector of size 6 which corresponds to the 6 toxicity classes to be predicted. Pre-trained models were used, and the models weights were unfrozen during training.
Multiple research have been done regarding features extraction in Transformers. Results presented in \citep{33} inspired our study and comparison. The paper shows that the concatenation of the four last layers from the encoder gives better results than using only the last layer.

\subparagraph{Compact Convolutional Transformer (CCT)} \citep{44} is a Transformer-based architecture for vision. The original paper shows that CCT can lead to good results on image and on text datasets with fewer parameters compared to Transformer based models. In previous research, some sought to use transformers on images: Vision Transformer (ViT) \citep{dosovitskiy2020vit}. The main idea of these models is to use the advantages of Transformers on images to extract information that cannot be brought out by convolutions. Unlike ViT, CCT combines convolutions and Transformer attention layers. CCT first uses a convolution tokenization on the image, while Vision Transformer uses patch-based tokenization. This layer applies a certain number of convolutions that produce a set of maps that are reshaped (flatten) and directly used by an optional positional embedding layer. The embedding is then fed to a series of Transformer encoder layers and pooled before being used by dense layers for classification.
We have recovered this type of transformer for our study to use it again on text. Since CCT works only on images, we used Glove pre-trained embedding to represent the sentences from the dataset as images. We padded the sentences to a fixed length and concatenated each embedded word to form a matrix representing a one channel image.
The training was done from scratch, and we used a pre-trained GloVe embedding enriched during the training.

 Global Vectors for Word Representation (GloVe) \citep{pennington-etal-2014-glove} is a model used to find word vectors. It uses a co-occurrence matrix to consider the global context of the words in the sentence. Semantic relationships between words can be extracted from the co-occurrence matrix.

To compare BERT-based models with other more traditional models, we trained a Bidirectional GRU and a Bidirectional LSTM from scratch. For each one, three layers of RNN and one layer of embedding unfreeze GloVe \citep{pennington-etal-2014-glove} were used.

Several ResNet \citep{45} with a depth of 44 and 56 were also trained from scratch. We used pre-trained GloVe embedding. In some sessions, we froze the embedding.



\section{Experiments}
\label{sec:experiments}

\subsection{Training}

All models are trained over three epochs, with a batch size of 32 examples, except for CCT, where we limit ourselves to 8 per batch due to the lack of VRAM. We use the AdamW optimizer.

We use the positive weighted Binary Cross Entropy (pwBCE) as a loss function. This loss function adds weights on the positive samples to consider them as much as the negatives. For the RoBERTa model, we use three other loss functions which are the Binary Cross Entropy (BCE), Focal Loss (FL) \citep{46}, and positive weighted Focal Loss (pwFL). The FL reduces the loss attributed to well-ranked examples and focuses on examples with poorly ranked classes, usually due to class imbalance. pwFL corresponds to the same trick as for pwBCE applied to FL.

To measure the model's performance, we use similar metrics that were used during the kaggle\foottotoref{civil-comments-data}: Macro AUROC, Macro F1 and Micro F1 with a threshold of 0.5, Precision and Recall.
To understand the model's complexity, we also measure the inference time per batch calculated on the test set. The average inference time per batch is computed from 6,000 batches during the inference phase on the test set.

As we can see in Table~\ref{tab:perfbias}, the hate speech detection models could make biased predictions for particular identities who are already the target of such abuse. To measure such unintended model
bias, we rely on the AUC-based metrics developed by \citet{23}. These include Subgroup AUC (Sub. AUC), Background Positive Subgroup Negative (BPSN) AUC, and Background Negative Subgroup Positive (BNSP) AUC. 

 \subparagraph{The sub. AUC} measures the AUROC for each identity using toxic and normal posts from the test set that mention the identity under consideration. A higher value means that a model is less likely to confuse the normal post that mentions the community with a toxic post that does not. 
 
 \subparagraph{The BPSN AUC} measures the AUROC for each identity, using normal posts that mention the identity and toxic posts that do not mention the identity under consideration. A higher value means that a model is less likely to confuse the normal post that mentions the community with a toxic post that does not. 
 
 \subparagraph{The BNSP AUC} measures the AUROC for each identity using toxic posts that mention the identity and normal posts that do not mention the identity under consideration from the test set. A higher value means that the model is less likely to confuse a toxic post that mentions the community with a normal post without one.\\
 
To combine these metrics across identities, we used the generalized mean (GM) or power mean with exponent $p$, which was already used by the Jigsaw/Conversation AI Team during a Kaggle competition\foottotoref{civil-comments-data}. So, we report the following three bias metrics for our comparison:

\begin{itemize}
    \item \textbf{GMB-Subgroup-AUC} is the GM for the Subgroup AUC 
    \item \textbf{GMB-BPSN-AUC} is the GM of the  BPSN AUC 
    \item \textbf{GMB-BNSP-AUC} is the GM of the  BNSP AUC
\end{itemize}

We restrict the evaluation to the test set only. By having this restriction, we can evaluate models in terms of bias reduction. Only identities with more than 500 examples in the test dataset will be included in the evaluation calculation.

\section{Results}
\label{sec:result}
\begin{table*}[htb]
\resizebox{1.0\linewidth}{!}{
\begin{tabular}{lllllllllll}
\toprule
 &  &  & \multicolumn{5}{c}{Performance} & \multicolumn{3}{c}{Bias} \\
\cmidrule(lr){4-8}
\cmidrule(lr){9-11}
 &  &  & AUROC & Macro F1 & Micro F1 & Precision & Recall & GMB Sub. & GMB BPSN & GMB BNSP \\
Model type & Id & Model name &  &  &  &  &  &  &  &  \\
\midrule
\multirow[c]{9}{*}{BERT} & 0 & AlBERT & {\cellcolor[HTML]{F1EBF5}} \color[HTML]{000000} 0.9790 & {\cellcolor[HTML]{03517E}} \color[HTML]{F1F1F1} 0.3463 & {\cellcolor[HTML]{045E93}} \color[HTML]{F1F1F1} 0.4786 & {\cellcolor[HTML]{045687}} \color[HTML]{F1F1F1} 0.3247 & {\cellcolor[HTML]{F2ECF5}} \color[HTML]{000000} 0.9104 & {\cellcolor[HTML]{DCDAEB}} \color[HTML]{000000} 0.8674 & {\cellcolor[HTML]{FEF6FA}} \color[HTML]{000000} 0.8998 & {\cellcolor[HTML]{2685BB}} \color[HTML]{F1F1F1} 0.9513 \\

 & 1 & BERTweet & {\cellcolor[HTML]{FEF6FB}} \color[HTML]{000000} 0.9816 & {\cellcolor[HTML]{0567A1}} \color[HTML]{F1F1F1} 0.3616 & {\cellcolor[HTML]{056FAF}} \color[HTML]{F1F1F1} 0.4928 & {\cellcolor[HTML]{04639B}} \color[HTML]{F1F1F1} 0.3363 & {\cellcolor[HTML]{FAF3F9}} \color[HTML]{000000} 0.9216 & {\cellcolor[HTML]{F9F2F8}} \color[HTML]{000000} 0.8780 & {\cellcolor[HTML]{F7F0F7}} \color[HTML]{000000} 0.8945 & {\cellcolor[HTML]{D9D8EA}} \color[HTML]{000000} 0.9603 \\

 & 2 & DistilBERT & {\cellcolor[HTML]{F8F1F8}} \color[HTML]{000000} 0.9804 & {\cellcolor[HTML]{3B92C1}} \color[HTML]{F1F1F1} 0.3879 & {\cellcolor[HTML]{3B92C1}} \color[HTML]{F1F1F1} 0.5115 & {\cellcolor[HTML]{167BB6}} \color[HTML]{F1F1F1} 0.3572 & {\cellcolor[HTML]{ECE7F2}} \color[HTML]{000000} 0.9001 & {\cellcolor[HTML]{F5EEF6}} \color[HTML]{000000} 0.8762 & {\cellcolor[HTML]{D3D4E7}} \color[HTML]{000000} 0.8740 & {\cellcolor[HTML]{FFF7FB}} \color[HTML]{000000} \bfseries 0.9644 \\

 & 3 & HateBERT & {\cellcolor[HTML]{F2ECF5}} \color[HTML]{000000} 0.9791 & {\cellcolor[HTML]{056FAE}} \color[HTML]{F1F1F1} 0.3679 & {\cellcolor[HTML]{05659F}} \color[HTML]{F1F1F1} 0.4844 & {\cellcolor[HTML]{045C90}} \color[HTML]{F1F1F1} 0.3292 & {\cellcolor[HTML]{F7F0F7}} \color[HTML]{000000} 0.9165 & {\cellcolor[HTML]{F1EBF4}} \color[HTML]{000000} 0.8744 & {\cellcolor[HTML]{F2ECF5}} \color[HTML]{000000} 0.8915 & {\cellcolor[HTML]{C6CCE3}} \color[HTML]{000000} 0.9589 \\

 & 4 & RoBERTa BCE & {\cellcolor[HTML]{FDF5FA}} \color[HTML]{000000} 0.9813 & {\cellcolor[HTML]{FFF7FB}} \color[HTML]{000000} \bfseries 0.4749 & {\cellcolor[HTML]{8EB3D5}} \color[HTML]{000000} 0.5359 & {\cellcolor[HTML]{589EC8}} \color[HTML]{F1F1F1} 0.3836 & {\cellcolor[HTML]{E0DEED}} \color[HTML]{000000} 0.8891 & {\cellcolor[HTML]{FEF6FA}} \color[HTML]{000000} 0.8800 & {\cellcolor[HTML]{F1EBF4}} \color[HTML]{000000} 0.8901 & {\cellcolor[HTML]{E8E4F0}} \color[HTML]{000000} 0.9616 \\

 & 5 & RoBERTa FL & {\cellcolor[HTML]{FFF7FB}} \color[HTML]{000000} \bfseries 0.9818 & {\cellcolor[HTML]{F4EEF6}} \color[HTML]{000000} 0.4648 & {\cellcolor[HTML]{BBC7E0}} \color[HTML]{000000} 0.5524 & {\cellcolor[HTML]{86B0D3}} \color[HTML]{000000} 0.4017 & {\cellcolor[HTML]{DBDAEB}} \color[HTML]{000000} 0.8839 & {\cellcolor[HTML]{FFF7FB}} \color[HTML]{000000} \bfseries 0.8807 & {\cellcolor[HTML]{FFF7FB}} \color[HTML]{000000} \bfseries 0.9010 & {\cellcolor[HTML]{D2D3E7}} \color[HTML]{000000} 0.9597 \\

 & 6 & RoBERTa pwBCE & {\cellcolor[HTML]{FBF3F9}} \color[HTML]{000000} 0.9809 & {\cellcolor[HTML]{045E93}} \color[HTML]{F1F1F1} 0.3541 & {\cellcolor[HTML]{05659F}} \color[HTML]{F1F1F1} 0.4845 & {\cellcolor[HTML]{045B8F}} \color[HTML]{F1F1F1} 0.3284 & {\cellcolor[HTML]{FBF4F9}} \color[HTML]{000000} 0.9232 & {\cellcolor[HTML]{F0EAF4}} \color[HTML]{000000} 0.8741 & {\cellcolor[HTML]{FBF4F9}} \color[HTML]{000000} 0.8982 & {\cellcolor[HTML]{ADC1DD}} \color[HTML]{000000} 0.9575 \\

 & 7 & RoBERTa pwFL & {\cellcolor[HTML]{FBF4F9}} \color[HTML]{000000} 0.9809 & {\cellcolor[HTML]{0566A0}} \color[HTML]{F1F1F1} 0.3612 & {\cellcolor[HTML]{0567A2}} \color[HTML]{F1F1F1} 0.4861 & {\cellcolor[HTML]{045D92}} \color[HTML]{F1F1F1} 0.3297 & {\cellcolor[HTML]{FDF5FA}} \color[HTML]{000000} 0.9254 & {\cellcolor[HTML]{EFE9F3}} \color[HTML]{000000} 0.8734 & {\cellcolor[HTML]{F3EDF5}} \color[HTML]{000000} 0.8920 & {\cellcolor[HTML]{D2D3E7}} \color[HTML]{000000} 0.9597 \\

 & 8 & XLM RoBERTa & {\cellcolor[HTML]{F2ECF5}} \color[HTML]{000000} 0.9790 & {\cellcolor[HTML]{023D60}} \color[HTML]{F1F1F1} 0.3368 & {\cellcolor[HTML]{034A74}} \color[HTML]{F1F1F1} 0.4680 & {\cellcolor[HTML]{03456C}} \color[HTML]{F1F1F1} 0.3135 & {\cellcolor[HTML]{FBF4F9}} \color[HTML]{000000} 0.9230 & {\cellcolor[HTML]{E1DFED}} \color[HTML]{000000} 0.8689 & {\cellcolor[HTML]{EBE6F2}} \color[HTML]{000000} 0.8859 & {\cellcolor[HTML]{B9C6E0}} \color[HTML]{000000} 0.9581 \\
\cline{1-11} 
CCT & 9 & CCT & {\cellcolor[HTML]{023858}} \color[HTML]{F1F1F1} 0.9505 & {\cellcolor[HTML]{034973}} \color[HTML]{F1F1F1} 0.3428 & {\cellcolor[HTML]{0569A4}} \color[HTML]{F1F1F1} 0.4874 & {\cellcolor[HTML]{0872B1}} \color[HTML]{F1F1F1} 0.3507 & {\cellcolor[HTML]{4A98C5}} \color[HTML]{F1F1F1} 0.7983 & {\cellcolor[HTML]{023858}} \color[HTML]{F1F1F1} 0.8133 & {\cellcolor[HTML]{3991C1}} \color[HTML]{F1F1F1} 0.8307 & {\cellcolor[HTML]{023858}} \color[HTML]{F1F1F1} 0.9447 \\
\cline{1-11} 
\multirow[c]{3}{*}{CNN} & 10 & Freeze GloVe ResNet44 & {\cellcolor[HTML]{034A74}} \color[HTML]{F1F1F1} 0.9526 & {\cellcolor[HTML]{9EBAD9}} \color[HTML]{000000} 0.4189 & {\cellcolor[HTML]{CACEE5}} \color[HTML]{000000} 0.5591 & {\cellcolor[HTML]{EEE8F3}} \color[HTML]{000000} 0.4631 & {\cellcolor[HTML]{023858}} \color[HTML]{F1F1F1} 0.7053 & {\cellcolor[HTML]{045A8D}} \color[HTML]{F1F1F1} 0.8219 & {\cellcolor[HTML]{023858}} \color[HTML]{F1F1F1} 0.7876 & {\cellcolor[HTML]{0A73B2}} \color[HTML]{F1F1F1} 0.9499 \\

 & 11 & Unfreeze GloVe ResNet44 & {\cellcolor[HTML]{71A8CE}} \color[HTML]{F1F1F1} 0.9660 & {\cellcolor[HTML]{EBE6F2}} \color[HTML]{000000} 0.4566 & {\cellcolor[HTML]{FFF7FB}} \color[HTML]{000000} \bfseries 0.5958 & {\cellcolor[HTML]{FFF7FB}} \color[HTML]{000000} \bfseries 0.4835 & {\cellcolor[HTML]{1E80B8}} \color[HTML]{F1F1F1} 0.7759 & {\cellcolor[HTML]{509AC6}} \color[HTML]{F1F1F1} 0.8421 & {\cellcolor[HTML]{86B0D3}} \color[HTML]{000000} 0.8493 & {\cellcolor[HTML]{65A3CB}} \color[HTML]{F1F1F1} 0.9540 \\

 & 12 & Unfreeze GloVe ResNet56 & {\cellcolor[HTML]{509AC6}} \color[HTML]{F1F1F1} 0.9639 & {\cellcolor[HTML]{1E80B8}} \color[HTML]{F1F1F1} 0.3778 & {\cellcolor[HTML]{358FC0}} \color[HTML]{F1F1F1} 0.5098 & {\cellcolor[HTML]{1C7FB8}} \color[HTML]{F1F1F1} 0.3604 & {\cellcolor[HTML]{CDD0E5}} \color[HTML]{000000} 0.8707 & {\cellcolor[HTML]{7EADD1}} \color[HTML]{F1F1F1} 0.8487 & {\cellcolor[HTML]{75A9CF}} \color[HTML]{F1F1F1} 0.8445 & {\cellcolor[HTML]{B4C4DF}} \color[HTML]{000000} 0.9579 \\
\cline{1-11} 
\multirow[c]{2}{*}{RNN} & 13 & BiGRU & {\cellcolor[HTML]{D7D6E9}} \color[HTML]{000000} 0.9748 & {\cellcolor[HTML]{045687}} \color[HTML]{F1F1F1} 0.3492 & {\cellcolor[HTML]{045A8D}} \color[HTML]{F1F1F1} 0.4762 & {\cellcolor[HTML]{045483}} \color[HTML]{F1F1F1} 0.3232 & {\cellcolor[HTML]{EEE9F3}} \color[HTML]{000000} 0.9036 & {\cellcolor[HTML]{AFC1DD}} \color[HTML]{000000} 0.8573 & {\cellcolor[HTML]{B0C2DE}} \color[HTML]{000000} 0.8616 & {\cellcolor[HTML]{D6D6E9}} \color[HTML]{000000} 0.9600 \\

 & 14 & BiLSTM & {\cellcolor[HTML]{DAD9EA}} \color[HTML]{000000} 0.9754 & {\cellcolor[HTML]{0569A5}} \color[HTML]{F1F1F1} 0.3638 & {\cellcolor[HTML]{328DBF}} \color[HTML]{F1F1F1} 0.5089 & {\cellcolor[HTML]{197DB7}} \color[HTML]{F1F1F1} 0.3586 & {\cellcolor[HTML]{D3D4E7}} \color[HTML]{000000} 0.8761 & {\cellcolor[HTML]{CED0E6}} \color[HTML]{000000} 0.8636 & {\cellcolor[HTML]{D7D6E9}} \color[HTML]{000000} 0.8758 & {\cellcolor[HTML]{A4BCDA}} \color[HTML]{000000} 0.9569 \\
\cline{1-11} 
XLNet & 15 & XLNet & {\cellcolor[HTML]{F7F0F7}} \color[HTML]{000000} 0.9800 & {\cellcolor[HTML]{023858}} \color[HTML]{F1F1F1} 0.3336 & {\cellcolor[HTML]{023858}} \color[HTML]{F1F1F1} 0.4586 & {\cellcolor[HTML]{023858}} \color[HTML]{F1F1F1} 0.3045 & {\cellcolor[HTML]{FFF7FB}} \color[HTML]{000000} \bfseries 0.9287 & {\cellcolor[HTML]{F0EAF4}} \color[HTML]{000000} 0.8738 & {\cellcolor[HTML]{E6E2EF}} \color[HTML]{000000} 0.8834 & {\cellcolor[HTML]{D2D3E7}} \color[HTML]{000000} 0.9597 \\
\bottomrule
\end{tabular}
}
\caption{Model performance results.}
\label{tab:perfbias}
\end{table*}

\begin{figure}[ht]
  \centering
  \includegraphics[width=.6\linewidth]{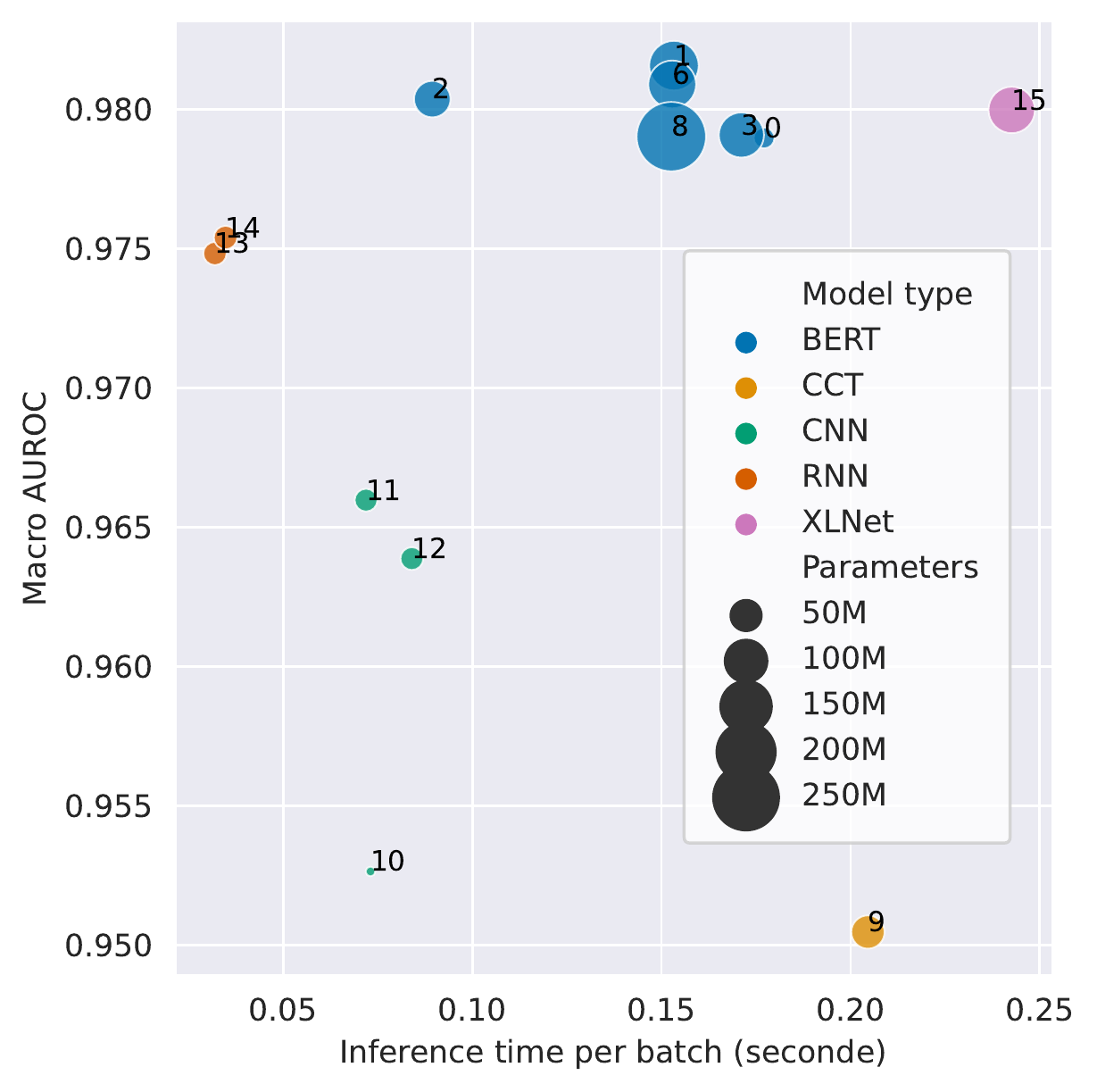}
  \caption{Model performance, depending on the inference time per batch and the number of trainable parameters. The number associated with each point corresponds to the model id in Table \ref{tab:perfbias}. All models have a batch of 32 samples, except CCT, which uses a batch of 8.}
  
  \label{graph:perftime}
\end{figure}

\begin{figure}[ht]
  \centering
  \includegraphics[width=\linewidth]{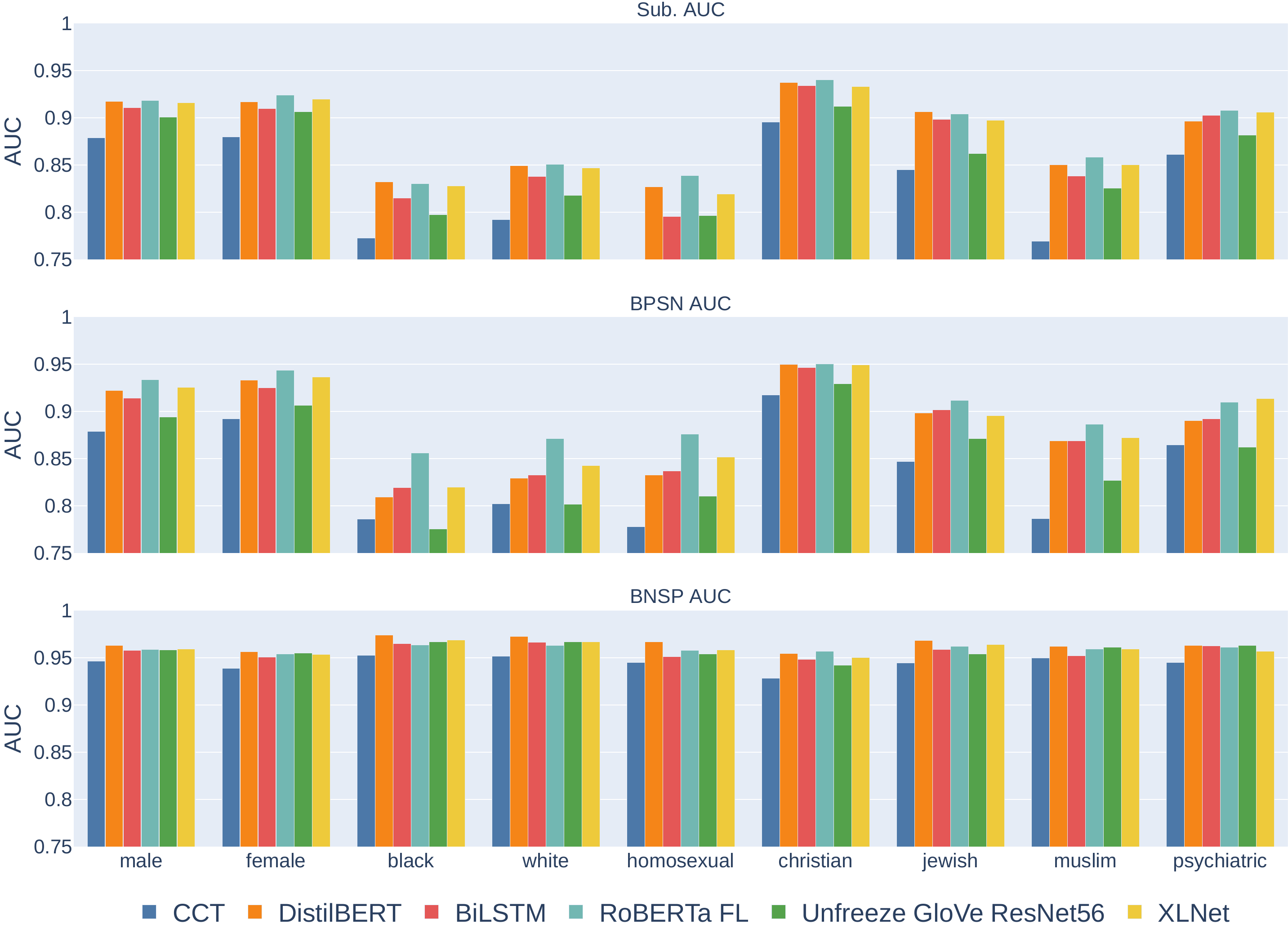}
  \caption{Community-wise results for each bias-metrics on the toxicity class. Only the most relevant models are shown here for the sake of readability. Thus, we have kept only the BERT, RNN, and CNN models with the best performance on AUROC or on the AUC bias metrics.}
  \label{graph:perfidentity}
\end{figure}

\subsection{Performances}

According to Figure \ref{graph:perftime} and Table \ref{tab:perfbias}, in general on the AUROC metric, BERT, RNN and XLNet have better scores than the others. As the comments are pretty short on average (27 tokens), RNN keep in memory a large part of the message to make a good prediction. We probably would have seen a more significant performance gap between BERT and RNN if the comments had been longer. RoBERTa with Focal Loss models offers the best performance on biases and AUROC. All BERT, regardless of size and optimizations, have similar performance. A DistilBERT or an AlBERT is as good as a HateBERT. Even if we notice that RoBERTa with the Focal Loss gives less biased results on the identity groups than DistillBERT. XLM RoBERTa does not differ from the others: the model has learned about other languages, but it does not give an advantage in detecting hateful comments.

Across the models, Recall is often very high, and Precision remains low. In other words, the models are more sensitive to hateful comments but generate more false positives. On the other hand, these same models detect many more true positive hate comments.

Within BERT, if we look at the RoBERTa with the different training losses (BCE, pwBCE, FL, pwFL), we see relatively close scores at the end. None of the tested training losses improve the learning of the models compared to a simple BCE.
We even note that the positive weights (pwBCE and pwFL) obtain worse F1 scores than the BCE or FL, but the Recall is 0.04 higher, and the accuracy is smaller by 0.1.

The Bi-GRU and Bi-LSTM have equivalent performance in terms of AUROC and F1 scores.


Nevertheless, we can show that all models have a little more difficulty classifying toxic comments and insults than explicit sexual comments.

\subsection{Bias}

Overall, if we look at the results presented in Table \ref{tab:perfbias}, we see that the models have a GMB BNSP greater than 0.95. In other words, the models have no difficulty differentiating between hateful comments targeting a community and generic comments (without targeting a particular identity).
On the contrary, we observe that the scores for GMB BPSN and GMB Sub are lower than those for GMB BNSP, often below 0.90. Thus, all the models present an association bias between identities and insults. They will tend to detect as being insults the positive comments about a community. But this bias depends on the model type.

From Table \ref{tab:perfbias}, we see that BERT and RNN models are generally less sensitive to this bias by having slightly higher GMB BPSN and GMB Sub. In contrast, convolution-based models such as CNN and CCT tend to be more sensitive to this association bias. CNN seek to capture patterns with convolutions.

From Figure \ref{graph:perfidentity}, all models score worse on average on BPSN and Sub. AUC for the \texttt{black}, \texttt{homosexual}, \texttt{muslim}, and \texttt{white} communities compared to the other communities.
For BPSN, this means that models have difficulty differentiating between insults that do not target identity and healthy comments about a community. Thus, these same models will tend to have more association bias and detect healthy comments about these communities as toxic.
For Sub. AUC, this means that when a comment targets an identity such as \texttt{black}, \texttt{gay}, \texttt{muslim}, or \texttt{white}, the models will have more difficulty distinguishing between hateful and non-hateful comments.

If we now look in more detail for each model and each identity, we notice again that \textit{RoBERTa with FL}, \textit{BiLSTM}, and \textit{XLNet} are less affected by that bias than \textit{Unfreeze GloVe ResNet56} and \textit{CCT}. There is even a difference of 0.05 on the Sub AUC for comments targeting communities such as \texttt{jewish} or \texttt{muslim} between \textit{RoBERTa with FL} and \textit{Unfreeze GloVe ResNet56}. Similarly, there is a difference of up to 0.1 on the BPSN AUC for the \texttt{black}, \texttt{homosexual} and \texttt{muslim} communities.
This shows that on these identities, which are particularly affected by hateful comments, the BERT, RNN, and XLNet models are less subject to association bias than the CNN and CCT.

\subsection{Inference time}

From Figure \ref{graph:perftime}, with performances quite close to the BERT type models, RNNs have an inference time, per batch, 5 to 8 times smaller than BERT or XLNet. As expected, DistilBERT, with the smallest inference time tested in our study, is 2 times higher than Bi-GRU and Bi-LSTM, even if the performance difference is 0.005 in AUROC.

The CNN ends up with an inference time shorter than most BERT and larger than the longest RNN tested, but with much lower performance than RNNs or BERT. With the same inference time per batch, DistilBERT does better.

We also notice that freezing the embedding does not decrease the inference time, but decreased the model's performance.

Finally, the CCT offers disappointing performances with a very long inference time per batch, especially when we know that we have reduced the batch size from 32 to 8 for this particular model.

\section{Conclusion}
\label{sec:conclusion}

All BERTs have similar performance regardless of the size, optimizations, or language used to pre-train the models. More broadly, BERT, RNN, and XLNet have almost similar performance. RNNs are much faster at inference than any of the BERT tested. RNNs remain a good compromise between performance and inference time for multi-label detection of hateful comments. RoBERTa with Focal Loss models offers the best performance on biases and AUROC. However, DistilBERT combines both good classification performance and a low inference time per batch. 

Even if the models are all affected by the bias of associating identities with toxicity, BERT, RNN, and XLNet are less sensitive to that than CNN and CCT.

\bibliographystyle{rnti}
\bibliography{biblio}



\Eng

\end{document}